%% file: acl2021.tex
\documentclass[11pt,a4paper]{article}
\usepackage[hyperref]{acl2021}
\usepackage{times}
\usepackage{latexsym}

\input{math_commands.tex}

\usepackage{algorithm}
\usepackage[noend]{algpseudocode}
\usepackage{amsthm}
\usepackage{booktabs}
\usepackage{caption}
\usepackage{graphicx}
\usepackage{hyperref}
\usepackage{url}
\usepackage{subcaption}
\usepackage{forest}
\usepackage{wrapfig}
\usepackage{multirow}
\forestset{
    nice empty nodes/.style={
        for tree={
            s sep=0.1em, 
            l sep=0.33em,
            inner ysep=0.4em, 
            inner xsep=0.05em,
            l=0,
            calign=midpoint,
            fit=tight,
            where n children=0{
               tier=word,
               minimum height=3em,
            }{},
            where n children=2{
               l-=0em,
            }{},
            parent anchor=south,
            child anchor=north,
            delay={if content={}{
                    inner sep=0pt,
                    edge path={\noexpand\path [\forestoption{edge}] 
                    			(!u.parent anchor) 
                               -- (.south)\forestoption{edge label};}
                }{}}
        },
    },
}
\usepackage{tikz}
\usepackage{qtree}
\usepackage{fancyhdr}

\theoremstyle{definition}
\newtheorem{definition}{Definition}[section]

\usepackage{microtype}

\aclfinalcopy 
\def\aclpaperid{663} 

\title{StructFormer: Joint Unsupervised Induction of Dependency and Constituency Structure from Masked Language Modeling}

\author{Yikang Shen\thanks{\quad  Corresponding author: \texttt{yikang.shn@gmail.com}. Work done while interning at Google Reseach.} \\
  Mila/Universit\'e de Montr\'eal \\ 
  \And
  Yi Tay \\
  Google Research \\
  \And
  Che Zheng \\
  Google Research \\
  \AND
  Dara Bahri \\
  Google Research \\
  \And
  Donald Metzler \\
  Google Research \\
  \And
  Aaron Courville \\
  Mila/Universit\'e de Montr\'eal \\
  }

\date{}

\begin{document}
\maketitle
\begin{abstract}
There are two major classes of natural language grammars --- the dependency grammar that models one-to-one correspondences between words and the constituency grammar that models the assembly of one or several corresponded words.
While previous unsupervised parsing methods mostly focus on only inducing one class of grammars, we introduce a novel model, StructFormer, that can simultaneously induce dependency and constituency structure. 
To achieve this, we propose a new parsing framework that can jointly generate a constituency tree and dependency graph. 
Then we integrate the induced dependency relations into the transformer, in a differentiable manner, through a novel dependency-constrained self-attention mechanism.
Experimental results show that our model can achieve strong results on unsupervised constituency parsing, unsupervised dependency parsing, and masked language modeling at the same time.
\footnote{Published as a conference paper at ACL 2021.}
\end{abstract}

\section{Introduction}
Human languages have a rich latent structure. This structure is multifaceted, with the two major classes of grammar being dependency and constituency structures. 
There has been an exciting breath of recent work  targeted at learning this structure in a data-driven unsupervised fashion \citep{klein2002generative, klein2005unsupervised, le2015unsupervised, shen2018ordered, kim2019compound}.
The core principle behind recent methods that induce structure from data is simple - provide an inductive bias that is conducive for structure to emerge as a byproduct of some self-supervised training, e.g., language modeling. 
To this end, a wide range of models have been proposed that are able to successfully learn grammar structures \citep{shen2018neural,shen2018ordered,wang2019tree, kim2019unsupervised,kim2019compound}. 
However, most of these works focus on inducing either constituency or dependency structures alone. 

In this paper, we make two important technical contributions. 
First, we introduce a new neural model, StructFormer, that is able to simultaneously induce \textbf{both} dependency structure and constituency structure. 
Specifically, our approach aims to unify latent structure induction of different types of grammar within the same framework.
Second, StructFormer is able to induce dependency structures from raw data in an end-to-end unsupervised fashion.
Most existing approaches induce dependency structures from other syntactic information like gold POS tags \citep{klein2004corpus,cohen2009shared,jiang2016unsupervised}.
Previous works, having trained from words alone, often requires additional information, like pre-trained word clustering \citep{spitkovsky2011unsupervised}, pre-trained word embedding \citep{he2018unsupervised}, acoustic cues \citep{pate2013unsupervised}, or annotated data from related languages \citep{cohen2011unsupervised}.

\begin{figure*}
    \centering
    \begin{subfigure}[h]{0.45\textwidth}
         \centering
         \includegraphics[width=\textwidth]{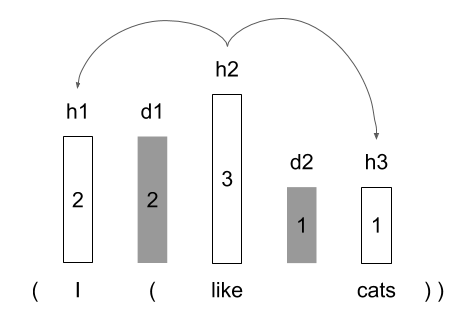}
         \caption{An example of Syntactic Distances $\distances$ (grey bars) and Syntactic Heights $\heights$ (white bars). 
         In this example, \texttt{like} is the parent (head) of constituent \texttt{(like cats)} and \texttt{(I like cats)}.
         }
         \label{fig:trees}
     \end{subfigure}
    \hfill
    \begin{subfigure}[h]{0.5\textwidth}
         \centering
         \includegraphics[width=\textwidth]{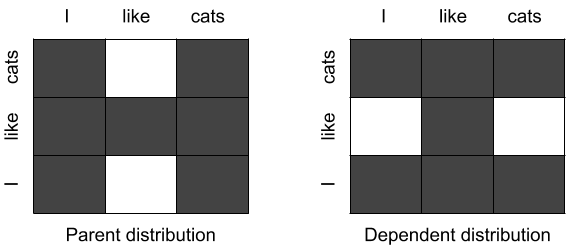}
         \caption{Two types of dependency relations. 
         The parent distribution allows each token to attend on its parent. 
         The dependent distribution allows each token to attend on its dependents. 
         For example the parent of \texttt{cats} is \texttt{like}.
         \texttt{Cats} and \texttt{I} are dependents of \texttt{like}
         Each attention head will receive a different weighted sum of these relations.
         }
         \label{fig:masks}
     \end{subfigure}
    \caption{An example of our parsing mechanism and dependency-constrained self-attention mechanism. 
    The parsing network first predicts the syntactic distance $\distances$ and syntactic height $\heights$ to represent the latent structure of the input sentence \texttt{I like cats}. 
    Then the parent and dependent relations are computed in a differentiable manner from $\distances$ and $\heights$.}
    \label{fig:examples}
\end{figure*}

We introduce a new inductive bias that enables the Transformer models to induce a directed dependency graph in a fully unsupervised manner.
To avoid the necessity of using grammar labels during training, we use a distance-based parsing mechanism.
The parsing mechanism predicts a sequence of Syntactic Distances $\distances$ \citep{shen2018straight} and a sequence of Syntactic Heights $\heights$ \citep{luo2019improving} to represent dependency graphs and constituency trees at the same time. 
Examples of $\heights$ and $\distances$ are illustrated in Figure \ref{fig:trees}.
Based on the syntactic distances ($\distances$) and syntactic heights ($\heights$), we provide a new dependency-constrained self-attention layer to replace the multi-head self-attention layer in standard transformer model. 
More specifically, the new attention head can only attend its parent (to avoid confusion with self-attention head, we use ``parent'' to denote ``head'' in dependency graph) or its dependents in the predicted dependency structure, through a weighted sum of relations shown in Figure \ref{fig:masks}.
In this way, we replace the complete graph in the standard transformer model with a differentiable directed dependency graph.
During the process of training on a downstream task (e.g. masked language model), the model will gradually converge to a reasonable dependency graph via gradient descent.

Incorporating the new parsing mechanism, the dependency-constrained self-attention, and the Transformer architecture, we introduce a new model named StructFormer. The proposed model can perform unsupervised dependency and constituency parsing at the same time, and can leverage the parsing results to achieve strong performance on masked language model tasks.

\section{Related Work}
Previous works on unsupervised dependency parsing are primarily based on the dependency model with valence (DMV) \citep{klein2004corpus} and its extension \citep{daume2009unsupervised, gillenwater2010sparsity}. 
To effectively learn the DMV model for better parsing accuracy, a variety of inductive biases and handcrafted features, such as correlations between parameters of grammar rules involving different part-of-speech (POS) tags, have been proposed to incorporate prior information into learning.
The most recent progress is the neural DMV model \citep{jiang2016unsupervised}, which uses a neural network model to predict the grammar rule probabilities based on the distributed representation of POS tags.
However, most existing unsupervised dependency parsing algorithms require the gold POS tags to ge provided as inputs. These gold POS tags are labeled by humans and can be potentially difficult (or prohibitively expensive) to obtain for large corpora.
\citet{spitkovsky2011unsupervised} proposed to overcome this problem with unsupervised word clustering that can dynamically assign tags to each word considering its context.
\citet{he2018unsupervised} overcame the problem by combining DMV model with invertible neural network to jointly model discrete syntactic structure and continuous word representations.

Unsupervised constituency parsing has recently received more attention. 
PRPN \citep{shen2018neural} and ON-LSTM \citep{shen2018ordered} induce tree structure by introducing an inductive bias to recurrent neural networks. 
PRPN proposes a parsing network to compute the syntactic distance of all word pairs, while a reading network
uses the syntactic structure to attend to relevant memories. 
ON-LSTM allows hidden neurons to learn long-term or short-term information by a novel gating mechanism and activation function. 
In URNNG \citep{kim2019unsupervised}, amortized variational inference was applied between a recurrent neural network grammar (RNNG) \citep{dyer2016recurrent} decoder and a tree structure inference network, which encourages the decoder to generate reasonable tree structures. 
DIORA \citep{drozdov2019unsupervised} proposed using inside-outside dynamic programming to compose latent representations from all possible binary trees. 
The representations of inside and outside passes from the same sentences are optimized to be close to each other.
The compound PCFG \citep{kim2019compound} achieves grammar induction by maximizing the marginal likelihood of the sentences which are generated by a probabilistic context-free grammar (PCFG).
Tree Transformer \citep{wang2019tree} adds extra locality constraints to the Transformer encoder's self-attention to encourage the attention heads to follow a tree structure such that each token can only attend on nearby neighbors in lower layers and gradually extend the attention field to further tokens when climbing to higher layers.
Neural L-PCFG \citep{zhu2020return} demonstrated that PCFG can benefit from modeling lexical dependencies. 
Similar to StructFormer, the Neural L-PCFG induces both constituents and dependencies within a single model.

Though large scale pre-trained models have dominated most natural language processing tasks, some recent work indicates that neural network models can see accuracy gains by leveraging syntactic information rather than ignoring it \citep{marcheggiani2017encoding, strubell2018linguistically}.
\citet{strubell2018linguistically} introduces syntactically-informed self-attention that force one attention head to attend on the syntactic governor of the input token.
\citet{omote2019dependency} and \citet{deguchi2019dependency} argue that dependency-informed self-attention can improve Transformer's performance on machine translation.
\citet{kuncoro2020syntactic} shows that syntactic biases help large scale pre-trained models, like BERT, to achieve better language understanding.

\section{Syntactic Distance and Height} \label{sec:distance}
In this section, we first reintroduce the concepts of syntactic distance and height, then discuss their relations in the context of StructFormer.

\subsection{Syntactic Distance}
Syntactic distance is proposed in \citet{shen2018straight} to quantify the process of splitting sentences into smaller constituents.
\begin{definition} \label{def:distance}
Let $\tree$ be a constituency tree for sentence $(w_1, ..., w_n)$.
The height of the lowest common ancestor for consecutive words $x_i$ and $x_{i+1}$ is $\Tilde{\tau}_i$. 
Syntactic distances $\distances=(\tau_1, ..., \tau_{n-1})$ are defined as a sequence of $n-1$ real scalars that share the same rank as $(\tilde{\tau}_1, ..., \tilde{\tau}_{n-1})$.
\end{definition}
In other words, each syntactic distance $d_i$ is associated with a split point $(i,i+1)$ and specify the relative order in which the sentence will be split into smaller components. 
Thus, any sequence of $n-1$ real values can unambiguously map to an unlabeled binary constituency tree with $n$ leaves through the Algorithm \ref{alg:distance2tree} \citep{shen2018straight}.
As \citet{shen2018ordered, shen2018neural, wang2019tree} pointed out, the syntactic distance reflects the information communication between constituents. 
More concretely, a large syntactic distance $\tau_i$ represents that short-term or local information should not be communicated between $(x_{\leq i})$ and $(x_{>i})$.
While cooperating with appropriate neural network architectures, we can leverage this feature to build unsupervised dependency parsing models.

\begin{minipage}{0.45\textwidth}
\begin{algorithm}[H]
    \centering
    \small
    \caption{Distance to binary constituency tree}\label{alg:distance2tree}
    \begin{algorithmic}[1]
    \Function{Constituent}{$\mathbf{w}$, $\mathbf{d}$}
    	\If {$\mathbf{d} = []$}
        	\State {$\tree \Leftarrow$ Leaf($\mathbf{w}$)}
        \Else
        	\State {$i \Leftarrow \mathrm{arg}\max_i (\mathbf{d})$}
            \State {$\mathrm{child}_l$ $\Leftarrow$ Constituent($\mathbf{w}_{\leq i}$, $\mathbf{d}_{<i}$)}
            \State {$\mathrm{child}_r$ $\Leftarrow$ Constituent($\mathbf{w}_{> i}$, $\mathbf{d}_{>i}$)}
            \State $\tree \Leftarrow \mathrm{Node}(\mathrm{child}_l, \mathrm{child}_r)$
        \EndIf
        \State \Return $\tree$
    \EndFunction
    \end{algorithmic}
\end{algorithm}
\end{minipage}

\begin{minipage}{0.45\textwidth}
\begin{algorithm}[H]
    \centering
    \small
    \caption{Converting binary constituency tree to dependency graph}\label{alg:height2tree}
    \begin{algorithmic}[1]
    \Function{Dependent}{$\tree$, $\heights$}
    	\If {$\tree = w$}
        	\State {$\dependent \Leftarrow [], \mathrm{parent} \Leftarrow w$}
        \Else 
            \State {$\mathrm{child}_l,\mathrm{child}_r \Leftarrow \tree$}
        	\State {$\dependent_l, \mathrm{parent}_l \Leftarrow \mathrm{Dependent}(\mathrm{child}_l, \heights)$}
        	\State {$\dependent_r, \mathrm{parent}_r \Leftarrow \mathrm{Dependent}(\mathrm{child}_r, \heights)$}
        	\State {$\dependent \Leftarrow \mathrm{Union}(\dependent_l, \dependent_r)$}
            \If {$\heights(\mathrm{parent}_l) > \heights(\mathrm{parent}_r)$}
                \State {$\dependent.\mathrm{add}(\mathrm{parent}_l \leftarrow \mathrm{parent}_r$)}
                \State {$\mathrm{parent}$ $\Leftarrow \mathrm{parent}_l$}
            \Else
                \State {$\dependent.\mathrm{add}(\mathrm{parent}_r \leftarrow \mathrm{parent}_l$)}
                \State {$\mathrm{parent}$ $\Leftarrow \mathrm{parent}_r$}
            \EndIf
        \EndIf
        \State \Return $\dependent, \mathrm{parent}$
    \EndFunction
    \end{algorithmic}
\end{algorithm}
\end{minipage}

\subsection{Syntactic Height} \label{sec:height}
Syntactic height is proposed in \citet{luo2019improving}, where it is used to capture the distance to the root node in a dependency graph. 
A word with high syntactic height means it is close to the root node. 
In this paper, to match the definition of syntactic distance, we redefine syntactic height as:
\begin{definition} \label{def:height}
Let $\dependent$ be a dependency graph for sentence $(w_1, ..., w_n)$. 
The height of a token $w_i$ in $\dependent$ is $\Tilde{\delta}_i$. 
The syntactic heights of $\dependent$ can be any sequence of $n$ real scalars $\heights=(\delta_1, ..., \delta_n)$ that share the same rank as $(\tilde{\delta}_1, ..., \tilde{\delta}_{n})$.
\end{definition}

Although the syntactic height is defined based on the dependency structure, we cannot rebuild the original dependency structure by syntactic heights alone, since there is no information about whether a token should be attached to the left side or the right side. 
However, given an unlabelled constituent tree, we can convert it into a dependency graph with the help of syntactic distance. 
The converting process is similar to the standard process of converting constituency treebank to dependency treebank \citep{gelbukh2005transforming}.
Instead of using the constituent labels and POS tags to identify the parent of each constituent, we simply assign the token with the largest syntactic height as the parent of each constituent.
The conversion algorithm is described in Algorithm \ref{alg:height2tree}.
In Appendix \ref{sec:jdp}, we also propose a joint algorithm, that takes $\distances$ and $\heights$ as inputs and jointly outputs a constituency tree and dependency graph. 

\begin{figure}[h]
    \centering
    \includegraphics[width=0.45\textwidth]{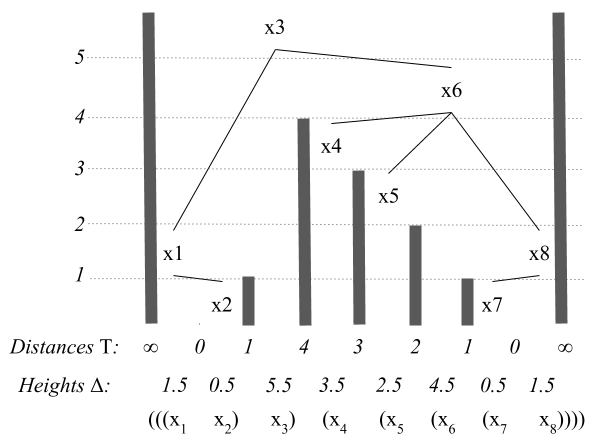}
    \caption{An example of $\distances$, $\heights$ and respective dependency graph $\dependent$.
    Solid lines represent dependency relations between tokens.
    StructFormer only allow tokens with dependency relation to attend on each other.
    }
    \label{fig:tree2}
\end{figure}

\subsection{The relation between Syntactic Distance and Height} \label{sec:relation}
As discussed previously, the syntactic distance controls information communication between the two sides of the split point.
The syntactic height quantifies the centrality of each token in the dependency graph.
A token with large syntactic height tends to have more long-term dependency relations to connect different parts of the sentence together.
In StructFormer, we quantify the syntactic distance and height on the same scale.
Given a split point $(i,i+1)$ and it's syntactic distance $\delta_i$, only tokens $x_j$ with $\tau_j > \delta_i$ can attend across the split point $(i,i+1)$.
Thus tokens with small syntactic height are limited to attend to nearby tokens.
Figure \ref{fig:tree2} provides an example of $\distances$, $\heights$ and respective dependency graph $\dependent$.

However, if the left and right boundary syntactic distance of a constituent $[l,r]$ are too large, all words in $[l,r]$ will be forced to only attend to other words in $[l,r]$. 
Their contextual embedding will not be able to encode the full context. 
To avoid this phenomena, we propose calibrating $\distances$ according to $\heights$ in Appendix \ref{app:calibration}

\section{StructFormer}
In this section, we present the StructFormer model. 
Figure \ref{fig:model} shows the architecture of StructFormer, which includes a parser network and a Transformer module.
The parser network predicts $\distances$ and $\heights$, then passes them to a set of differentiable functions to generate dependency distributions.
The Transformer module takes these distributions and the sentence as input to computes a contextual embedding for each position.
The StructFormer can be trained in an end-to-end fashion on a Masked Language Model task.
In this setting, the gradient back propagates through the relation distributions into the parser.

\subsection{Parsing Network}

\begin{figure}
    \centering
    \begin{subfigure}[b]{0.27\textwidth}
         \centering
         \includegraphics[width=\textwidth]{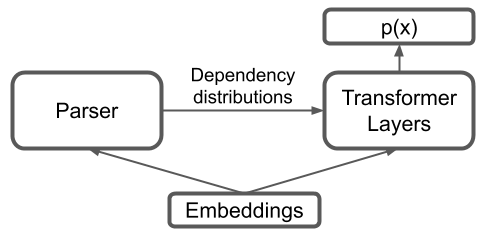}
         \caption{Model Architecture}
         \label{fig:model}
     \end{subfigure}
    \begin{subfigure}[b]{0.2\textwidth}
         \centering
         \includegraphics[width=\textwidth]{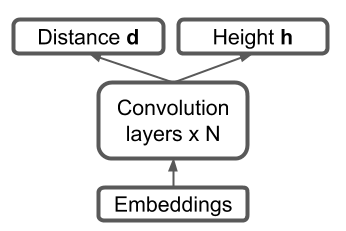}
         \caption{Parsing Network}
         \label{fig:parser}
     \end{subfigure}
    \caption{The Architecture of StructFormer. 
    The parser takes shared word embeddings as input, outputs syntactic distances $\distances$, syntactic heights $\heights$, and dependency distributions between tokens.
    The transformer layers take word embeddings and dependency distributions as input, output contextualized embeddings for input words.}
    \label{fig:architecture}
\end{figure}

As shown in Figure \ref{fig:parser}, the parsing network takes word embeddings as input and feeds them into several convolution layers:
\begin{eqnarray}
    s_{l,i} = \tanh \left( \mathrm{Conv} \left(s_{l-1, i - W}, ..., s_{l-1, i + W} \right) \right)
\end{eqnarray}
where $s_{l,i}$ is the output of $l$-th layer at $i$-th position, $s_{0, i}$ is the input embedding of token $w_i$, and $2W + 1$ is the convolution kernel size.

Given the output of the convolution stack $s_{N,i}$, we parameterize the syntactic distance $\distances$ as:
\begin{eqnarray}
    \tau_i = \left\{ 
    \begin{array}{r}
    \mathbf{W}^{\tau}_1 \tanh \left( \mathbf{W}^{\tau}_2
    \left[ \begin{matrix} 
        s_{N, i} \\
        s_{N, i+1}
    \end{matrix} \right] \right) , \\ 
    1 \leq i \leq n - 1 \\
    \infty, \quad i = 0 \quad \mathrm{or} \quad i = n
    \end{array}
    \right.
\end{eqnarray}
where $\tau_i$ is the contextualized distance for the $i$-th split point between token $w_i$ and $w_{i+1}$.
The syntactic height $\heights$ is parameterized in a similar way:
\begin{eqnarray}
    \delta_i = \mathbf{W}^{\delta}_1 \tanh \left( \mathbf{W}^{\delta}_2 s_{N, i} + b^{\delta}_2 \right) + b^{\delta}_1
\end{eqnarray}

\subsection{Estimate the Dependency Distribution}

Given $\distances$ and $\heights$, we now explain how to estimate the probability $p(x_j|x_i)$ such that the $j$-th token is the parent of the $i$-th token.
The first step is identifying the smallest legal constituent $\constituent(x_i)$, that contains $x_i$ and $x_i$ is not $\constituent(x_i)$'s parent. 
The second step is identifying the parent of the constituent $x_j = \parent(\constituent(x_i))$.
Given the discussion in section \ref{sec:height}, the parent of $\constituent(x_i)$ must be the parent of $x_i$.
Thus, the two-stages of identifying the parent of $x_i$ can be formulated as:
\begin{equation}
    \dependent(x_i) = \parent(\constituent(x_i))
\end{equation}
In StructFormer, $\constituent(x_i)$ is represented as constituent $[l,r]$, where $l$ is the starting index ($l \leq i$) of $\constituent(x_i)$ and $r$ is the ending index ($r \geq i$) of $\constituent(x_i)$. 

In a dependency graph, $x_i$ is only connected to its parent and dependents.
This means that $x_i$ does not have direct connection to the outside of $\constituent(x_i)$.
In other words, $\constituent(x_i) = [l, r]$ is the smallest constituent that satisfies:
\begin{eqnarray}
    \delta_i < \tau_{l-1}, & \delta_i < \tau_{r}
\end{eqnarray}
where $\tau_{l-1}$ is the first $\tau_{<i}$ that is larger then $\delta_i$ while looking backward, and $\tau_{r}$ is the first $\tau_{\geq i}$ that is larger then $\delta_i$ while looking forward.
For example, in Figure \ref{fig:tree2}, $\delta_4 = 3.5$, $\tau_3 = 4 > \delta_4$ and $\tau_8 = \infty > \delta_4$, thus $\constituent(x_4) = [4, 8]$.
To make this process differentiable, we define $\tau_k$ as a real value and $\delta_i$ as a probability distribution $p(\tilde{\delta}_i)$.
For the simplicity and efficiency of computation, we directly parameterize the cumulative distribution function $p(\tilde{\delta}_i > \tau_{k})$ with sigmoid function:
\begin{eqnarray}
    p(\tilde{\delta}_i > \tau_{k}) = \sigma ((\delta_i - \tau_{k})/\mu_1)
\end{eqnarray}
where $\sigma$ is the sigmoid function, $\delta_i$ is the mean of distribution $p(\tilde{\delta}_i)$ and $\mu_1$ is a learnable temperature term.
Thus the probability that the $l$-th ($l < i$) token is inside $\constituent(x_i)$ is equal to the probability that $\tilde{\delta}_i$ is larger then the maximum distance $\tau$ between $l$ and $i$:
\begin{eqnarray}
    p(l \in \constituent(x_i)) &=& p(\tilde{\delta}_i > \max(\tau_{i-1}, ..., \tau_{l})) \\
    &=& \sigma ((\delta_i - \max(\tau_{l}, ..., \tau_{i-1}))/\mu) \nonumber 
\end{eqnarray}
Then we can compute the probability distribution for $l$:
\begin{eqnarray}
    p(l|i) &=& 
    p(l \in \constituent(x_i)) - p(l - 1 \in \constituent(x_i)) \nonumber \\
     &=& \sigma ((\delta_i - \max(\tau_{l}, ..., \tau_{i-1}))/\mu) - \nonumber \\
     & & \sigma ((\delta_i - \max(\tau_{l-1}, ..., \tau_{i-1}))/\mu)
\end{eqnarray}
Similarly, we can compute the probability distribution for $r$:
\begin{eqnarray}
    p(r|i) &=& \sigma ((\delta_i - \max(\tau_{i}, ..., \tau_{r-1}))/\mu) - \nonumber \\ 
    & & \sigma ((\delta_i - \max(\tau_{i}, ..., \tau_{r}))/\mu)
\end{eqnarray}
The probability distribution for $[l, r] = \constituent(x_i)$ can be computed as:
\begin{eqnarray}
    p_{\constituent}([l,r] | i) &=& \left\{ \begin{matrix}
        p(l|i)p(r|i), & l \leq i \leq r \\
        0, & \mathrm{otherwise}
    \end{matrix} \right.
\end{eqnarray}
The second step is to identify the parent of $[l,r]$. 
For any constituent $[l,r]$, we choose the $j = \mathrm{argmax}_{k \in [l,r]}(\delta_k)$ as the parent of $[l, r]$.
In the previous example, given constituent $[4, 8]$, the maximum syntactic height is $\delta_6 = 4.5$, thus $\parent([4,8]) = x_6$.
We use softmax function to parameterize the probability $p_{\parent}(j|[l,r])$:
\begin{eqnarray}
    p_{\parent}(j|[l,r]) = \left\{ \begin{matrix}
        \frac{\exp(h_j / \mu_2)}{\sum_{l \leq k \leq r} \exp(h_k / \mu_2)}, & l \leq t \leq r \\
        0, & \mathrm{otherwise}
    \end{matrix} \right.
\end{eqnarray}
Given probability $p(j|[l,r])$ and $p([l,r]|i)$, we can compute the probability that $x_j$ is the parent of $x_i$:
\begin{equation}
    p_{\dependent}(j|i) = \left\{ \begin{matrix}
        \sum_{[l,r]} p_{\parent}(j|[l,r]) p_{\constituent}([l,r]|i), & i \neq j \\
        0, & i = j
    \end{matrix} \right.
\end{equation}

\subsection{Dependency-Constrained Multi-head Self-Attention}
The multi-head self-attention in the transformer can be seen as a information propagation mechanism on the complete graph $\mathbf{G} = (X, E)$, where the set of vertices $X$ contains all $n$ tokens in the sentence, and the set of edges $E$ contains all possible word pairs $(x_i, x_j)$.
StructFormer replace the complete graph $\mathbf{G}$ with a soft dependency graph $\dependent = (X, A)$, where $A$ is the matrix of $n \times n$ probabilities. $A_{ij} = p_{\dependent} (j|i)$ is the probability of the $j$-th token depending on the $i$-th token.
The reason that we called it a directed edge is that each specific head is only allow to propagate information either from parent to dependent or from from dependent to parent.
To do so, structformer associate each attention head with a probability distribution over parent or dependent relation.
\begin{eqnarray}
    p_{\mathrm{parent}} &=& \frac{\exp(w_{\mathrm{parent}})}{\exp(w_{\mathrm{parent}}) + \exp(w_{\mathrm{dep}})} \\
    p_{\mathrm{dep}} &=& \frac{\exp(w_{\mathrm{dep}})}{\exp(w_{\mathrm{parent}}) + \exp(w_{\mathrm{dep}})}
\end{eqnarray}
where $w_{\mathrm{parent}}$ and $w_{\mathrm{dep}}$ are learnable parameters that associated with each attention head, $p_{\mathrm{parent}}$ is the probability that this head will propagate information from parent to dependent, vice versa.
The model will learn to assign this association from the downstream task via gradient descent.
Then we can compute the probability that information can be propagated from node $j$ to node $i$ via this head:
\begin{eqnarray}
    p_{i,j} = p_{\mathrm{parent}} p_{\dependent}(j|i) + p_{\mathrm{dep}} p_{\dependent}(i|j)
\end{eqnarray}
However, \citet{htut2019attention} pointed out that different heads tend to associate with different type of universal dependency relations (including \texttt{nsubj}, \texttt{obj}, \texttt{advmod}, etc), but there is no generalist head can that work with all different relations.
To accommodate this observation, we compute a individual probability for each head and pair of tokens $(x_i, x_j)$:
\begin{equation}
    q_{i,j} = \mathrm{sigmoid} \left( \frac{QK^T}{\sqrt{d_k}} \right)
\end{equation}
where $Q$ and $K$ are query and key matrix in a standard transformer model and $d_k$ is the dimension of attention head.
The equation is inspired by the scaled dot-product attention in transformer. 
We replace the original softmax function with a sigmoid function, so $q_{i,j}$ became an independent probability that indicates whether $x_i$ should attend on $x_j$ through the current attention head.
In the end, we propose to replace transformer's scaled dot-product attention with our dependency-constrained self-attention:
\begin{equation}
    \mathrm{Attention}(Q_i,K_j,V_j,\dependent) = p_{i,j} q_{i,j} V_j
\end{equation}

\section{Experiments}
We evaluate the proposed model on three tasks: Masked Language Modeling, Unsupervised Constituency Parsing and Unsupervised Dependency Parsing.

Our implementation of StructFormer is close to the original Transformer encoder \citep{vaswani2017attention}. 
Except that we put the layer normalization in front of each layer, similar to the T5 model \citep{raffel2019exploring}.
We found that this modification allows the model to converges faster.
For all experiments, we set the number of layers $L=8$, the embedding size and hidden size to be $d_{model} = 512$, the number of self-attention heads $h=8$, the feed-forward size $d_{ff}=2048$, dropout rate as $0.1$,
and the number of convolution layers in the parsing network as $L_{p} = 3$.

\subsection{Masked Language Model}
Masked Language Modeling (MLM) has been widely used as a pretraining object for larger-scale pretraining models.
In BERT \citep{devlin2018bert} and RoBERTa \citep{liu2019roberta}, authors found that MLM perplexities on held-out evaluation set have a positive correlation with the end-task performance.
We trained and evaluated our model on 2 different datasets: the Penn TreeBank (PTB) and BLLIP.
In our MLM experiments, each token, including $<unk>$ token, has an independent chance to be replaced by a mask token \texttt{<mask>}.
The training and evaluation object for Masked Language Model is to predict the replaced tokens.
The performance of MLM is evaluated by measuring perplexity on masked words. 

{\bf PTB} is a standard dataset for language modeling \citep{mikolov2012statistical} and unsupervised constituency parsing \citep{shen2018ordered, kim2019compound}.
Following the setting proposed in \citet{shen2018ordered}, we use \citet{mikolov2012statistical}'s prepossessing process, which removes all punctuations, and replaces low frequency tokens with \texttt{<unk>}. The preprocessing results in a vocabulary size of 10001 (including \texttt{<unk>}, \texttt{<pad>} and \texttt{<mask>}).
For PTB, we use a 30\% mask rate.

{\bf BLLIP} is a large Penn Treebank-style parsed corpus of approximately 24 million sentences.
We train and evaluate StructFormer on three splits of BLLIP: BLLIP-XS (40k sentences, 1M tokens), BLLIP-SM (200K sentences, 5M tokens), and BLLIP-MD (600K sentences, 14M tokens). 
They are obtained by randomly sampling sections from BLLIP 1987-89 Corpus Release 1.
All models are tested on a shared held-out test set (20k sentences, 500k tokens).
Following the settings provided in \citep{hu2020systematic}, we use subword-level vocabulary extracted from the GPT-2 pre-trained model rather than the BLLIP training corpora. 
For BLLIP, we use a 15\% mask rate.

\begin{table}[t]
    \centering
    \small
    \begin{tabular}{l c c c c}
    \toprule
        \multirow{2}{*}{Model} & \multirow{2}{*}{PTB} & BLLIP & BLLIP & BLLIP \\
         & & -XS & -SM & -MD \\
    \midrule
        Transformer & 64.05 & 93.90 & 19.92 & 14.31 \\
        StructFormer & 60.94 & 57.28 & 18.70 & 13.70 \\
    \bottomrule
    \end{tabular}
    \caption{Masked Language Model perplexities on different datasets.}
    \label{tab:bllip}
    \vspace{-0.3cm}
\end{table}

The masked language model results are shown in Table \ref{tab:bllip}. 
StructFormer consistently outperforms our Transformer baseline.
This result aligns with previous observations that linguistically informed self-attention can help Transformers achieve stronger performance. 
We also observe that StructFormer converges much faster than the standard Transformer model.

\subsection{Unsupervised Constituency Parsing}
The unsupervised constituency parsing task compares the latent tree structure induced by the model with those annotated by human experts.
We use the Algorithm \ref{alg:distance2tree} to predict the constituency trees from $\distances$ predicted by StructFormer.
Following the experiment settings proposed in \citet{shen2018ordered},
we take the model trained on PTB dataset and evaluate it on WSJ test set.
The WSJ test set is section 23 of WSJ corpus, it contains 2416 human expert labeled sentences.
Punctuation is ignored during the evaluation.

\begin{table}[h]
    \centering
    \small
    \begin{tabular}{c c}
    \toprule
        Methods & UF1  \\
        \midrule
        RANDOM & 21.6 \\
        LBRANCH & 9.0 \\
        RBRANCH & 39.8 \\
        \midrule
        PRPN \citep{shen2018neural} & 37.4 (0.3) \\
        ON-LSTM \citep{shen2018ordered} & 47.7 (1.5) \\
        Tree-T \citep{wang2019tree} & 49.5 \\
        URNNG \citep{kim2019unsupervised} & 52.4 \\
        C-PCFG \citep{kim2019compound} & 55.2 \\
        Neural L-PCFGs \citep{zhu2020return} & 55.31 \\
        StructFormer & 54.0 (0.3) \\
        \bottomrule
    \end{tabular}
    \caption{Unsupervised constituency parsing tesults. 
    * results are from \citet{kim2020pre}.
    UF1 stands for Unlabeled F1.}
    \label{tab:constituency}
\end{table}

\begin{table}[h]
    \centering
    \small
    \begin{tabular}{r c c c c c}
        \toprule
         & PRPN & ON & C-PCFG & Tree-T & Ours \\
        \midrule
        SBAR & 50.0\% & 52.5\% & \bf 56.1\% & 36.4\% & 48.7\% \\
        NP & 59.2\% & 64.5\% & \bf 74.7\% & 67.6\% & 72.1\% \\
        VP & \bf 46.7\% & 41.0\% & 41.7\% & 38.5\% & 43.0\% \\
        PP & 57.2\% & 54.4\% & 68.8\% & 52.3\% & \bf 74.1\% \\
        ADJP & 44.3\% & 38.1\% & 40.4\% & 24.7\% & \bf 51.9\% \\
        ADVP & 32.8\% & 31.6\% & 52.5\% & 55.1\% & \bf 69.5\% \\
        \bottomrule
    \end{tabular}
    \caption{Fraction of ground truth constituents that were predicted as a constituent by the models broken down by label (i.e. label recall)}
    \label{tab:phrase}
\end{table}

Table \ref{tab:constituency} shows that our model achieves strong results on unsupervised constituency parsing.
While the C-PCFG \citep{kim2019compound} achieve a stronger parsing performance with its strong linguistic constraints (e.g. a finite set of production rules), StructFormer may have a border domain of application.
For example, it can replace the standard transformer encoder in most of the popular large-scale pre-trained language models (e.g. BERT and ReBERTa) and transformer based machine translation models.
Different from the transformer-based Tree-T \citep{wang2019tree}, we did not directly use constituents to restrict the self-attention receptive field.
But StructFormer achieves a stronger constituency parsing performance.
This result may suggest that dependency relations are more suitable for grammar induction in transformer-based models.
Table \ref{tab:phrase} shows that our model achieves strong accuracy while predicting Noun Phrase (NP), Preposition Phrase (PP), Adjective Phrase (ADJP), and Adverb Phrase (ADVP).

\subsection{Unsupervised Dependency Parsing}

\begin{table*}[t]
    \centering
    \small
    \begin{tabular}{c c c c c c c}
    \toprule
        Relations & MLM & Constituency & \multicolumn{2}{c}{Stanford}  & \multicolumn{2}{c}{Conll} \\
         & PPL & UF1 & UAS & UUAS  & UAS & UUAS \\
        \midrule
        parent+dep & 60.9 (1.0) & 54.0 (0.3) & 46.2 (0.4) & 61.6 (0.4) & 36.2 (0.1) & 56.3 (0.2) \\
        parent & 63.0 (1.2) & 40.2 (3.5) & 32.4 (5.6) & 49.1 (5.7) & 30.0 (3.7) & 50.0 (5.3) \\
        dep & 63.2 (0.6) & 51.8 (2.4) & 15.2 (18.2) & 41.6 (16.8) & 20.2 (12.2) & 44.7 (13.9) \\
        \bottomrule
    \end{tabular}
    \caption{The performance of StructFormer with different combinations of attention masks. 
    UAS stands for Unlabeled Attachment Score.
    UUAS stands for Undirected Unlabeled Attachment Score.}
    \label{tab:attend_masks}
\end{table*}

The unsupervised dependency parsing evaluation compares the induced dependency relations with those in the reference dependency graph.
The most common metric is the Unlabeled Attachment Score (UAS), which measures the percentage that a token is correctly attached to its parent in the reference tree.
Another widely used metric for unsupervised dependency parsing is Undirected Unlabeled Attachment Score (UUAS) measures the percentage that the reference undirected and unlabeled connections are recovered by the induced tree.
Similar to the unsupervised constituency parsing, we take the model trained on PTB dataset and evaluate it on WSJ test set (section 23).
For the WSJ test set, reference dependency graphs are converted from its human-annotated constituency trees.
However, there are two different sets of rules for the conversion: the Stanford dependencies and the CoNLL dependencies.
While Stanford dependencies are used as reference dependencies in previous unsupervised parsing papers, we noticed that our model sometimes output dependency structures that are closer to the CoNLL dependencies.
Therefore, we report UAS and UUAS for both Stanford and CoNLL dependencies.
Following the setting of previous papers \citep{jiang2016unsupervised}, we ignored the punctuation during evaluation.
To obtain the dependency relation from our model, we compute the argmax for dependency distribution:
\begin{equation}
    k = \mathrm{argmax}_{j \neq i} p_{\dependent}(j|i)
\end{equation}
and assign the $k$-th token as the parent of $i$-th token.

\begin{table}[t]
    \centering
    \small
    \begin{tabular}{c c}
    \toprule
        Methods & UAS \\
        \midrule
        \multicolumn{2}{l}{\small \emph{w/o gold POS tags}} \\
        DMV \citep{klein2004corpus} &  35.8 \\
        E-DMV \citep{headden2009improving} & 38.2 \\
        UR-A E-DMV \citep{tu2012unambiguity} & 46.1 \\
        CS* \citep{spitkovsky2013breaking} & 64.4* \\
        Neural E-DMV \citep{jiang2016unsupervised} & 42.7 \\
        Gaussian DMV \citep{he2018unsupervised} & 43.1 (1.2) \\
        INP \citep{he2018unsupervised} & 47.9 (1.2) \\
        Neural L-PCFGs \citep{zhu2020return} & 40.5 (2.9) \\
        StructFormer & 46.2 (0.4) \\
        \midrule
        \multicolumn{2}{l}{\small \emph{w/ gold POS tags (for reference only)}} \\
        DMV \citep{klein2004corpus} & 39.7 \\
        UR-A E-DMV \citep{tu2012unambiguity} & 57.0 \\
        MaxEnc \citep{le2015unsupervised} & 65.8 \\
        Neural E-DMV \citep{jiang2016unsupervised} & 57.6 \\
        CRFAE \citep{cai2017crf} & 55.7 \\
        L-NDMV$^\dagger$ \citep{han2017dependency} & 63.2 \\
        \bottomrule
    \end{tabular}
    \caption{Dependency Parsing Results on WSJ testset. 
    Starred entries (*) benefit from additional punctuation-based constraints.
    Daggered entries ($^\dagger$) benefit from larger additional training data.
    Baseline results are from \citet{he2018unsupervised}.
    }
    \label{tab:dependency}
\end{table}

Table \ref{tab:dependency} shows that our model achieves competitive dependency parsing performance while comparing to other models that do not require gold POS tags.
While most of the baseline models still rely on some kind of latent POS tags or pre-trained word embeddings, StructFormer can be seen as an easy-to-use alternative that works in an end-to-end fashion.
Table \ref{tab:attend_masks} shows that our model recovers 61.6\% of undirected dependency relations.
Given the strong performances on both dependency parsing and masked language modeling, we believe that the dependency graph schema could be a viable substitute for the complete graph schema used in the standard transformer.
Appendix \ref{app:dep_distribution} provides examples of parent distribution.



Since our model uses a mixture of the relation probability distribution for each self-attention head, we also studied how different combinations of relations affect the performance of our model.
Table \ref{tab:attend_masks} shows that the model can achieve the best performance while using both parent and dependent relations.
The model suffers more on dependency parsing if the parent relation is removed.
And if the dependent relationship is removed, the model will suffer more on the constituency parsing.
Appendix \ref{app:rel_weight} shows the weight for parent and dependent relations learnt from MLM tasks. 
It's interesting to observe that Structformer tends to focus on the parent relations in the first layer, and start to use both relations from the second layer.

\section{Conclusion}
In this paper, we introduce a novel dependency and constituency joint parsing framework.
Based on the framework, we propose StructFormer, a new unsupervised parsing algorithm that does unsupervised dependency and constituency parsing at the same time.
We also introduced a novel dependency-constrained self-attention mechanism that allows each attention head to focus on a specific mixture of dependency relations.
This brings Transformers closer to modeling a directed dependency graph.
The experiments show promising results that StructFormer can induce meaningful dependency and constituency structures and achieve better performance on masked language model tasks.
This research provides a new path to build more linguistic bias into a pre-trained language model.

\bibliographystyle{acl_natbib}
\bibliography{anthology,acl2021}

\include{appendix}

\end{document}

%% file: math_commands.tex
\newcommand{\distances}{\mathrm{T}}
\newcommand{\heights}{\Delta}
\newcommand{\constituent}{\mathbf{C}}

\newcommand{\parent}{\mathbf{Pr}}
\newcommand{\dependent}{\mathbf{D}}
\newcommand{\tree}{\mathbf{T}}

\usepackage{amsmath,amsfonts,bm}

\def\eqref#1{equation~\ref{#1}}

\def\1{\bm{1}}

\DeclareMathAlphabet{\mathsfit}{\encodingdefault}{\sfdefault}{m}{sl}
\SetMathAlphabet{\mathsfit}{bold}{\encodingdefault}{\sfdefault}{bx}{n}



%% file: appendix.tex
\appendix
\section{Appendix}
\subsection{Joint Dependency and Constituency Parsing} \label{sec:jdp}
\begin{algorithm}[h]
\small
\caption{The joint dependency and constituency parsing algorithm. 
Inputs are a sequence of words $\mathbf{w}$, syntactic distances $\mathbf{d}$, syntactic heights $\mathbf{h}$. 
Outputs are a binary constituency tree $\tree$, a dependency graph $\dependent$ that is represented as a set of dependency relations, the $\mathrm{parent}$ of dependency graph $\dependent$, and the syntactic $\mathrm{height}$ of $\mathrm{parent}$.}
\label{alg:jdp}
\begin{algorithmic}[1]
\Function{BuildTree}{$\mathbf{w},\mathbf{d},\mathbf{h}$}
	\If {$\mathbf{d} = []$ and $\mathbf{w} = [w]$ and $\mathbf{h} = [h]$}
    	\State {$\tree \Leftarrow \mathrm{Leaf}(w)$, $\dependent \Leftarrow []$, $\mathrm{parent} \Leftarrow w$, $\mathrm{height}$ $\Leftarrow h$}
    \Else
    	\State {$i \Leftarrow \arg\max (\mathbf{d})$}
    	\State {$\tree_l, \dependent_l, \mathrm{parent}_l, \mathrm{height}_l$ $\Leftarrow$ BuildTree($\mathbf{d}_{< i}, \mathbf{w}_{\leq i}, \mathbf{h}_{\leq i}$)}
    	\State {$\tree_r, \dependent_r, \mathrm{parent}_r, \mathrm{height}_r$ $\Leftarrow$ BuildTree($\mathbf{d}_{> i}, \mathbf{w}_{> i}, \mathbf{h}_{> i}$)}
    	\State {$\tree \Leftarrow \mathrm{Node}(\mathrm{child}_l \Leftarrow \tree_l, \mathrm{child}_r \Leftarrow \tree_r)$}
    	\State {$\dependent$ $\Leftarrow \mathrm{Union}(\dependent_l$, $\dependent_r$)}
        \If {$\mathrm{height}_l > \mathrm{height}_r$}
            \State {$\dependent.\mathrm{add}(\mathrm{parent}_l \leftarrow \mathrm{parent}_r$)}
            \State {$\mathrm{parent}$ $\Leftarrow \mathrm{parent}_l$, $\mathrm{height}$ $\Leftarrow \mathrm{height}_l$}
        \Else
            \State {$\dependent.\mathrm{add}(\mathrm{parent}_r \leftarrow \mathrm{parent}_l$)}
            \State {$\mathrm{parent}$ $\Leftarrow \mathrm{parent}_r$, $\mathrm{height}$ $\Leftarrow \mathrm{height}_r$}
        \EndIf
    \EndIf
    \State \Return $\tree$, $\dependent$, $\mathrm{parent}$, $\mathrm{height}$
\EndFunction
\end{algorithmic}
\end{algorithm} 

\subsection{Calibrating the Syntactic Distance and Height} \label{app:calibration}
In Section \ref{sec:relation}, we explained the relation between $\heights$ and $\distances$, that if $\delta_i < \tau_j$, the $i$-th word won't be able to attend beyond the $j$-th split point.
However, in a specific case, the constraint will isolate a constituent $[l,r]$ from the rest of the sentence.
If $\tau_{l-1}$ and $\tau_{r}$ are larger then all height ${\delta_{l,...,r}}$ in the constituent, then all words in $[l, r]$ won't be able to attend on the outside of the constituent.
This phenomenon will prevent their output contextual embedding from encoding the full context.
To avoid this phenomenon, we propose to calibrate the syntactic distance $\distances$ according to the syntactic height $\heights$.
First, we compute the maximum syntactic height for each constituent:
\begin{eqnarray}
    \delta_{[l,r]} = \max \left(\delta_l, ..., \delta_r \right), \quad l < r
\end{eqnarray}
Then we compute the minimum difference between $\delta_{[l,r]}$ and $[l,r]$'s left and right boundary distance. 
Since we only care about constituents that the boundary distance is larger than its maximum height, we use a ReLU activation function to keep only the positive values:
\begin{equation}
    \epsilon_{[l,r]} = \mathrm{ReLU} \left( \min \left( \tau_{l-1} - \delta_{[l,r]}, \tau_r - \delta_{[l,r]} \right) \right)
\end{equation}
To make sure all constituent are not isolated and maintain the rank of $\distances$, we subtract all $\distances$ by the maximum of $\epsilon$:
\begin{equation}
    \hat{\delta_i} = \delta_i - \max_{\{[l,r]\} / [1,n]} \left( \epsilon_{[l,r]} \right)
\end{equation}

\onecolumn
\newpage
\subsection{Dependency Relation Weights for Self-attention Heads}
\label{app:rel_weight}
\begin{figure}[H]
    \centering
    \begin{subfigure}[h]{0.8\textwidth}
        \centering
        \includegraphics[width=1\textwidth]{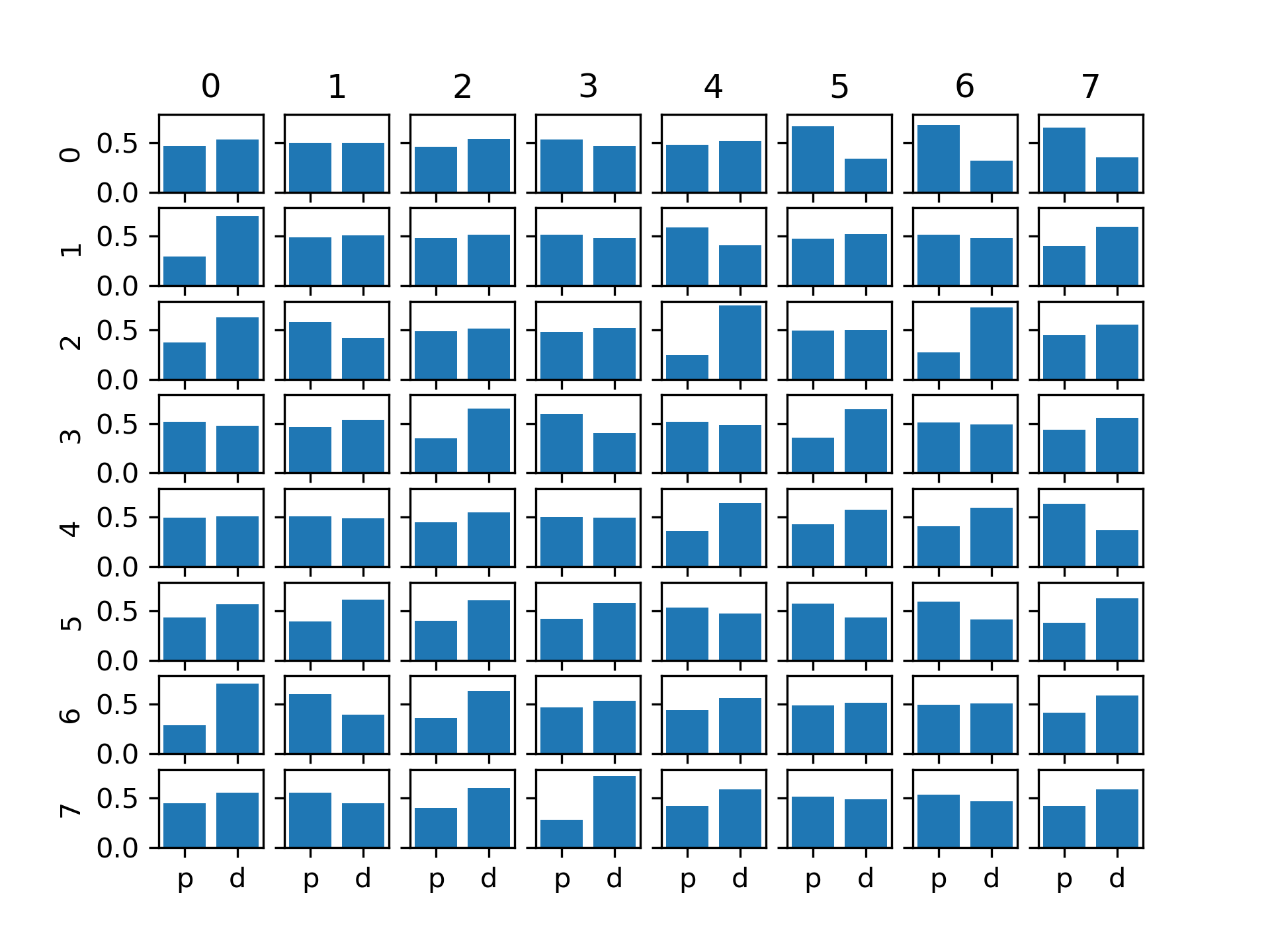}
        \caption{Dependency relation weights learnt on PTB}
    \end{subfigure}
    
    \begin{subfigure}[h]{0.8\textwidth}
        \centering
        \includegraphics[width=1\textwidth]{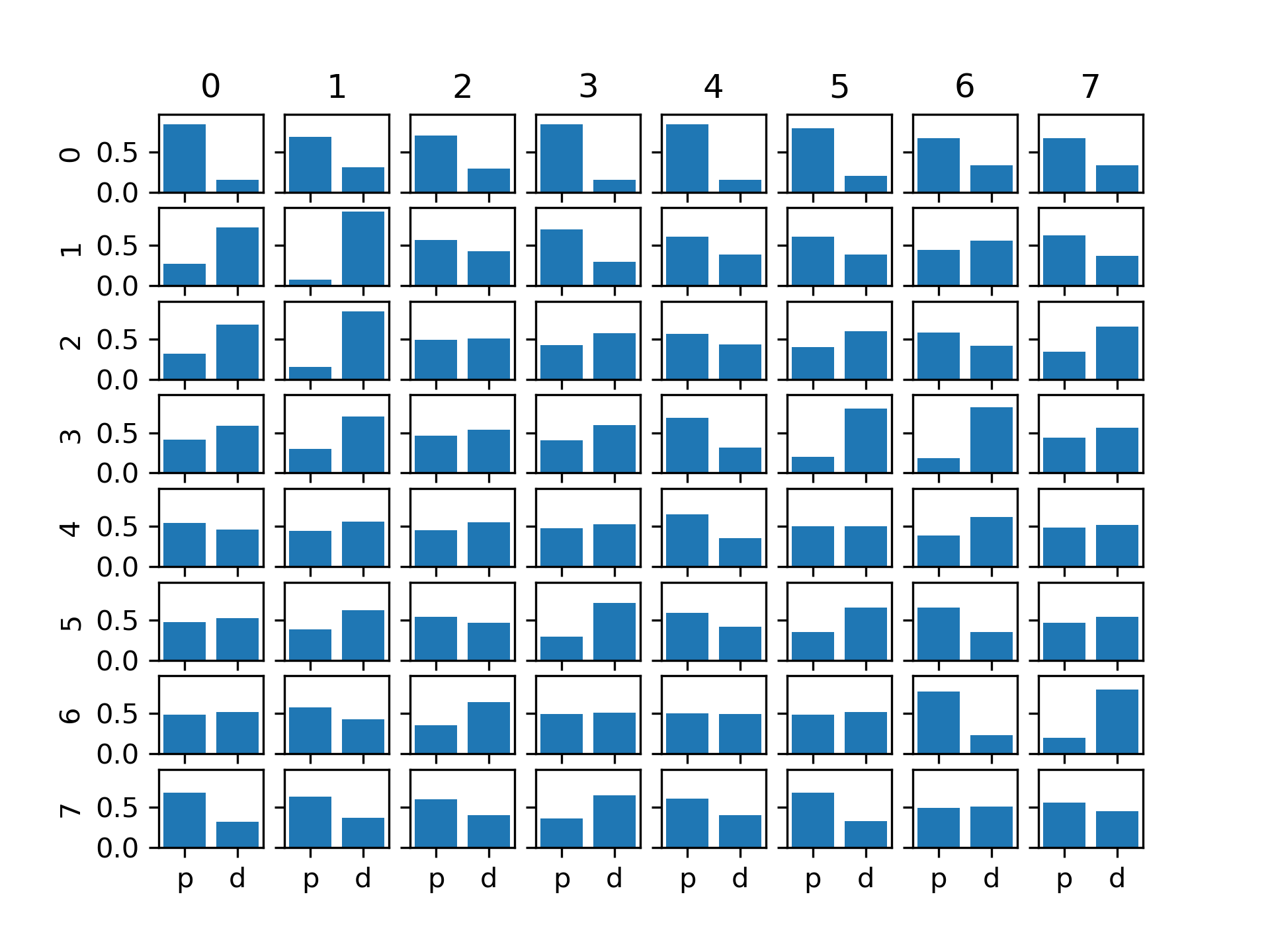}
        \caption{Dependency relation weights learnt on BLLIP-SM}
    \end{subfigure}
    
    \caption{Dependency relation weights learnt on different datasets. 
        Row $i$ constains relation weights for all attention heads in the $i$-th transformer layer.
        \texttt{p} represents the parent relation.
        \texttt{d} represents the dependent relation.
        We observe a clearer preference for each attention head in the model trained on BLLIP-SM. 
        This probably due to BLLIP-SM has signficantly more training data.
        It's also interesting to notice that the first layer tend to focus on parent relations.
        }
    \label{fig:mask_weights}
\end{figure}

\newpage
\subsection{Dependency Distribution Examples}
\label{app:dep_distribution}
\begin{figure}[H]
    \centering
    \begin{subfigure}[h]{0.8\textwidth}
        \centering
        \includegraphics[width=1\textwidth]{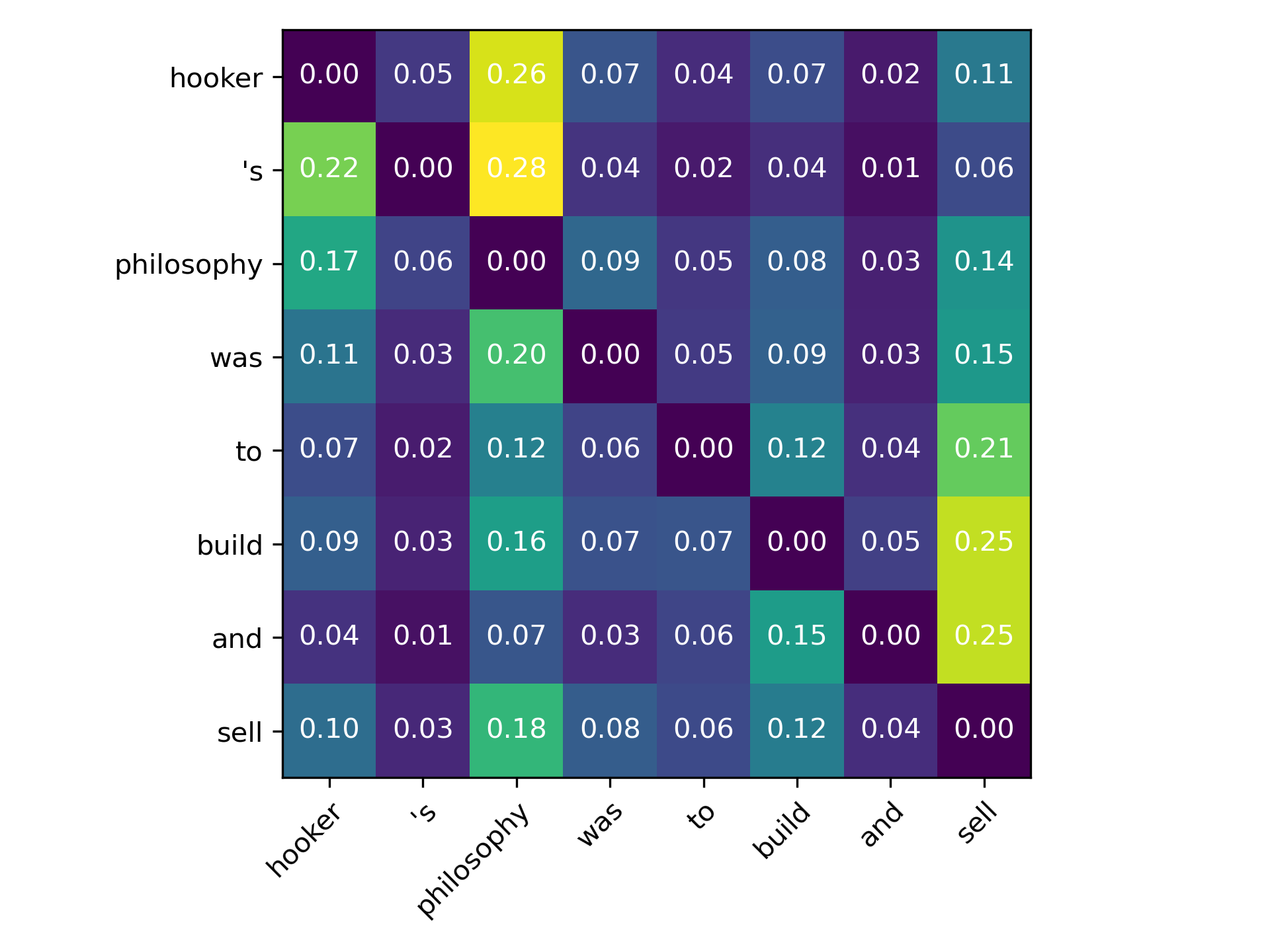}
        \caption{}
    \end{subfigure}
    
    \begin{subfigure}[h]{0.8\textwidth}
        \centering
        \includegraphics[width=1\textwidth]{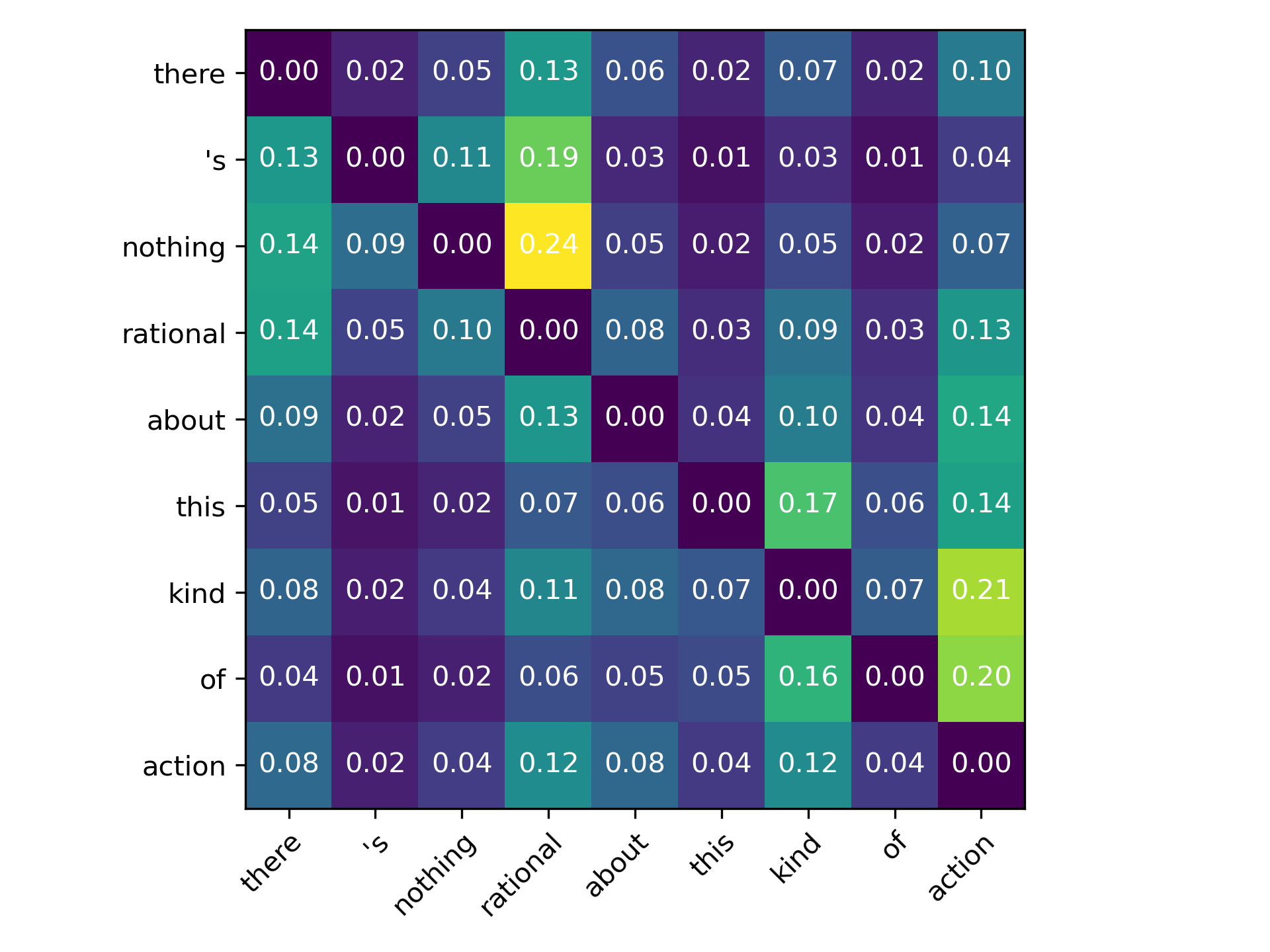}
        \caption{}
    \end{subfigure}
    
    \caption{Dependency distribution examples from WSJ test set. 
        Each row is the parent distribution for the respective word.
        The sum of each distribution may not equal to 1.
        Against our intuition, the distribution is not very sharp.
        This is partially due to the ambiguous nature of the dependency graph.
        As we previously discussed, at least two styles of dependency rules (Conll and Stanford) exist. 
        And without extra constraint or supervision, the model seems trying to model both of them at the same time.
        One interesting future work will be finding an inductive bias that can encourage the model to converge to a specific style of dependency graph.
        }
    \label{fig:dependency_distribution}
\end{figure}

\newpage
\subsection{The Performance of StructFormer with different mask rates}
\begin{table}[H]
    \centering
    \small
    \begin{tabular}{c c c c c c c}
    \toprule
        Mask rate & MLM & Constituency & \multicolumn{2}{c}{Stanford}  & \multicolumn{2}{c}{Conll} \\
         & PPL & UF1 & UAS & UUAS  & UAS & UUAS \\
        \midrule
        0.1 & 45.3 (1.2) & 51.45 (2.7) & 31.4 (11.9) & 51.2 (8.1) & 32.3 (5.2) & 52.4 (4.5) \\
        0.2 & 50.4 (1.3) & 54.0 (0.6) & 37.4 (12.6) & 55.6 (8.8) & 33.0 (5.7) & 53.5 (4.7) \\
        0.3 & 60.9 (1.0) & 54.0 (0.3) & 46.2 (0.4) & 61.6 (0.4) & 36.2 (0.1) & 56.3 (0.2) \\
        0.4 & 76.9 (1.2) & 53.5 (1.5) & 34.0 (10.3) & 52.0 (7.4) & 29.5 (5.4) & 50.6 (4.1) \\
        0.5 & 100.3 (1.4) & 53.2 (0.9) & 36.3 (9.8) & 53.6 (6.8) & 30.6 (4.2) & 51.3 (3.2) \\
        \bottomrule
    \end{tabular}
    \caption{The performance of StructFormer on PTB dataset with different mask rates. 
    Dependency parsing is especially affected by the masks.
    Mask rate 0.3 provides the best and the most stable performance.}
    \label{tab:masks_rates}
\end{table}